\documentclass[runningheads]{llncs}
\usepackage[T1]{fontenc}
\usepackage{lmodern}
\usepackage{graphicx}
\usepackage{epstopdf}
\usepackage{subcaption}
\usepackage{hyperref}

\usepackage{amsmath} % Add this line to include the amsmath package
\usepackage{amsfonts}
\usepackage{algorithm}
\usepackage{algpseudocode}
\usepackage{lipsum,adjustbox}
\usepackage{multicol}
\usepackage[square, sort, comma, numbers]{natbib}

\begin{document}

%

% \title{\textbf{Graph Neural Alchemist:}\\An innovative fully modular architecture for classifying time series as graphs}

\title{\textbf{Graph Neural Alchemist:}\\An innovative fully modular architecture for time series-to-graph classification}
\titlerunning{Graph Neural Alchemist for Time Series Classification}
\author{Paulo Coelho\inst{1},
Raul Araju \inst{1},
Luís Ramos \inst{1},
Samir Saliba \inst{1} \and
Renato Vimieiro \inst{1}
}
\authorrunning{Coelho. Paulo et al.}

\institute{Institute of Exact Sciences -- Computer Science Department\\Federal University of Minas Gerais (UFMG), Belo Horizonte, MG, Brazil
\email{\{paulohdscoelho,raularaju,luisfeliperamos,samirsaliba,rvimieiro\} @dcc.ufmg.br}\\
\url{https://dcc.ufmg.br/}}
\maketitle              % typeset the header of the contribution

\begin{abstract}
This paper introduces a novel Graph Neural Network (GNN) architecture for time series classification, based on visibility graph representations. Traditional time series classification methods often struggle with high computational complexity and inadequate capture of spatio-temporal dynamics. By representing time series as visibility graphs, it is possible to encode both spatial and temporal dependencies inherent to time series data, while being computationally efficient. Our architecture is fully modular, enabling flexible experimentation with different models and representations. We employ directed visibility graphs encoded with in-degree and PageRank features to improve the representation of time series, ensuring efficient computation while enhancing the model's ability to capture long-range dependencies in the data. We show the robustness and generalization capability of the proposed architecture across a diverse set of classification tasks and against a traditional model. Our work represents a significant advancement in the application of GNNs for time series analysis, offering a powerful and flexible framework for future research and practical implementations.
\keywords{Deep Learning \and Time Series and Classification \and Graph Neural Networks \and Visibility Graphs.}
\end{abstract}
\section{Introduction}~\label{sec:introduction}

Time Series are all around us: from financial markets and weather forecasting to physiological signals such as electrocardiograms (ECGs), the ubiquitous nature of time series has sparked increasing interest in applying Deep Neural Network (DNN) models for analysis. Those models, however, rely heavily on feature extraction techniques and often fail to capture the intrinsic and complex spatio-temporal dynamics inherent in time series data, besides being computationally expensive and lacking interpretability. To address that, time series-to-graph representation has emerged as a powerful tool to represent spatial relationships within data while encoding graph-metrics that capture temporal dependences between its elements, and it has the following advantages: graphs offer a more interpretable and easily modeled structure; and, DNN models designed for graph learning tend to be more computationally efficient than traditional ones.

Among series-to-graph representations, Visibility Graph (VG) stands out for its capability to effectively capture spatial relationships inherent in time series data. The topology of a VG is intrinsically tied to the nature of the original time series; periodic series translate into periodic graphs, while random series into random graphs~\citep{lacasa2008time}. Also, from the VG we can extract graph-metrics to encapsulate temporal dynamics within data. By employing Graph Neural Networks (GNNs) models on VGs we could learn the hidden spatio-temporal patterns, translating the problem of Time Series Classification into a Graph Classification problem, a highly explainable and computationally efficient supervised learning task when compared to traditional time series classification.

In this paper, we propose the \textbf{Graph Neural Alchemist} (GNA), a architecture for time series classification comprised by: an innovative series-to-graph representation by Visibility Graphs encoded with local and global metrics in-Degree and PageRank, a GNN model with a readout layer to learn over the graph representation and a modified Multilayer Perceptron to perform the Graph Classification of time series. Our proposed architecture is fully modular, i.e., we designed it in a way that the series-to-graph representation in the first module, or the GNN model for instance, can be easily changed depending on the subsequent task and the data domain at hand.

We begin by discussing related works, followed by a theoretical background, and then present our proposed GNA implementation using visibility graphs to represent time series and GNN to learn their \textit{spatio-temporal dynamics}. We formalize the architecture and demonstrate that it performs well in time series classification in comparision with traditional time series classification models, achieving promising results on benchmark datasets from the UCR Time Series Classification Archive.

\section{Related Works}~\label{sec:related_works}

In the Great Time Series Classification Bakeoff~\citep{bagnall2017great}, the authors conducted an extensive review of time series classification algorithms based on similarity, interval, and shapelet representations, ranking the best ones in terms of mean accuracy on the UCR Archive datasets.One such traditional approach, ROCKET (RandOm Convolutional KErnel Transform)~\cite{dempster2020rocket}, has shown remarkable performance across a wide range of time series classification tasks. ROCKET utilizes a large number of random convolutional kernels to transform the time series data, followed by a simple linear classifier. This approach has been noted for its efficiency and effectiveness, particularly in balanced datasets.

Despite their effectiveness, these models exhibit high time complexity and rely heavily on feature extraction to represent and learn from data, failing to directly capture the spatio-temporal dynamics inherent in time series. To address this issue, time series-to-graph representation techniques have been proposed to capture temporal dependencies and structure of the data~\citep{wang2020deep, sahili2023spatio, 10026346}.

Existing architectures comprise various frameworks and algorithms, each employing different strategies for time series representation, such as transition, proximity or visibility graphs, and diverse GNN models to learn hidden features.~\citet{jin2307survey} surveyed advancements in time series-to-graph representation for graph classification, presenting the series-as-graph and series-as-node architectures Time2Graph and SimTSC. While it presents important series-to-graph strategies, the survey didn't address visibility graphs techniques.

The Time2Graph~\citep{cheng2020time2graph} captures spatio-temporal relations by representing time series as an evolution graph, where nodes are time-aware shapelets extracted from the time series, and edges are based on the temporal transition probabilities between these shapelets. This approach uses DeepWalk for classification, while Time2Graph+~\citep{cheng2021time2graphplus} employs a Graph Attention Network (GAT) to learn the graph structure and transition probabilities.

~\citet{zha2022similarityaware} proposed SimTSC, a modular architecture to encode similarity between time series, where each serie in the training set is mapped as a node in a proximity graph, with edges based on pairwise Dynamic Time Warping (DTW) similarity between series. Node features are learned by a ResNet backbone and then fed into a Graph Convolutional Network (GCN) model followed by a 1-NN classifier for node classification. Although these approaches show promising results, they are parameter-dependent and rely on the extraction of shapelets or pairwise similarities, which can be computationally expensive, and do not properly encode local and global features from the data.

~\citet{lacasa2008time} introduced Visibility Graphs (VG), a parameter-free algorithm to time series-to-graph representation, where each node corresponds to a data point, and edges are drawn between nodes that are mutually visible. Although easy to understand and computationally efficient~\citep{lan2015fast}, vanilla VGs lack the ability to encode temporal information directly on graphs.

~\citet{sbcas2020} proposed a method for cardiac arrhythmia classification that represents ECG time series as directed VGs encoded with in-degree and PageRank as node features to encode temporal information on the graph, testing it with several GNN architectures for classification. While innovative, the study benchmark the model only on one type of time series data, and the architecture lacks modularity.

~\citet{coelho2024} employed visibility graph representations of multivariate 12-lead ECG time series and proposed a framework to calculate a graph diversity measure~\citep{carpi2019assessing} for ranking leads (i.e., ECG channels) based on their contribution to the series' overall diversity. This measure was then used to select the top 3 most diverse leads to validate if there were improvments in the classification performance against a Convolutional model. The study is a founding step on time series-to-graph representation based on visibility graphs.

Our proposed GNA architecture advances the aforementioned works by implementing directed visibility graphs encoded with in-degree and PageRank features to time series-to-graph representation and employing a GNN model to learn the spatio-temporal dynamics of the data. The proposed architecture is modular, with each step of the task (representation, learning, and classification) being self-contained in a separate module, allowing for easy experimentation with different models, datasets, and tasks.

\section{Theoretical Background}~\label{sec:theoretical_background}

In this section we present the theoretical background needed to understand our work. We start by discussing Visibility Graph, which is used to time series-to-graph representation, followed by a brief explanation on the GNNs model and the \textit{Neural Message Passing Framework} as defined by~\citet{hamilton2020graph}.

\subsection{Visibility Graph}~\label{subsec:visibility_graph}

Proposed by~\citet{lacasa2008time}, Natural Visibility Graph (NVG) is a method of time series-to-graph representation that preserves the structure and dynamics of the original series. Formally, the NVG defines a graph $\mathcal{G} = (\mathcal{V}, \space \mathcal{E})$, where vertices represent values of the time series and edges are defined based on a visibility criterion between those values. 

Informally, we can put the general idea of NVG as follows: given a time series $T$ of dimension $m$, its values $\{t_1, t_2, \dots t_m\}$ are projected onto a Cartesian plane, where the temporal dimension of the series corresponds to the $x$-axis and its absolute value to the $y$-axis. More formally, two arbitrary data values $A(t_a, y_a)$ and $B(t_b, y_b)$ are mutually visible if there is no point $C(t_c, y_c)$ between $A$ and $B$ that is above the line connecting $A$ and $B$, as defined by the equation:

\begin{equation}~\label{eq:visibility_criterion}
    y_c < y_b + (y_a - y_b) \frac{t_b - t_c}{t_b - t_a},
\end{equation}
where $t_a < t_c < t_b$.

It is worth mentioning that resulting NVG is: 
\begin{itemize}
    \item \textbf{Undirected:} The visibility between two points is symmetric;
    \item \textbf{Connected:} Adjacent points are always visible to each other;
    \item \textbf{Invariant under affine transformations:} The Visibility criterion is invariant under rescaling of both $x$ and $y$'s axis and under horizontal and vertical translations.
\end{itemize}

~\citet{lan2015fast} proposed an efficient methodology that computes VGs in time $\text{O}(n \log n)$, where $n = |V|$, making it more practical and efficient for large-scale time series data.

\subsection{Graph Neural Networks}~\label{subsec:graph_neural_netowks}

GNNs work directly on graph structure and compute vertex representations by aggregating neighboring node embeddings. The input for a GNN is a set of node feature vectors, which are real-valued characteristics of each node in the graph. GNN models aim to learn low-dimensional representations of the nodes that summarize their graph position and the structure of their local graph neighborhood into a latent space $\textbf{z} \in \mathbb{R}^d$, where $d$ is the dimension of the feature vector~\citep{hamilton2018representation}.

GNNs can be understood by an Encoder-Decoder perspective, where the Encoder aggregates and updates information for every node in the graph based on its local neighborhood; learning a hidden embedding representation of the node features that is used by the Decoder to reconstruct information about each node's neighborhood in the original graph to perform downstream tasks such as node or graph classification~\citep{hamilton2020graph}.

The \textit{Neural Message Passing Framework} is fundamental to understand how the GNN model learns hidden node embeddings, and can be expressed as: given a graph $\mathcal{G} (\mathcal{V}, \space \mathcal{E})$ and the node feature embeddings $\{h_u \in \mathbb{R}^d\}$ where $d$ is the feature dimension, for each node $u \in \mathcal{V}$ of $\mathcal{G}$, information about the current node $u$ and its neighbourhood $\mathcal{N}(u)$ are iteratively exchanged and updated across the edges that connects them, producing an \textit{hidden embedding} $\textbf{h}_u^{(k)}$ that summarizes the information of the node and its neighbourhood at the $k$-th iteration. The message passing framework can be formalized as:

\begin{equation}~\label{eq:message_passing}
    \begin{aligned}
        \textbf{h}_u^{(k+1)} &= UPDATE^{(k)}(\textbf{h}_u^{(k)}, \space AGGREGATE^{(k)}(\{\textbf{h}_v^{(k)}, \forall v \in \mathcal{N}(u) \})) \\
        &= UPDATE^{(k)} (\textbf{h}_u^{(k)}, \textbf{m}_{\mathcal{N}(u)}^{(k)}),
    \end{aligned}
\end{equation} 

where $UPDATE$ and $AGGREGATE$ are differentiable arbitrary functions such as Neural Networks, $\textbf{m}_{\mathcal{N}(u)}$ is the ``message'' aggreagated from the neighbourhood graph $\mathcal{N}(u)$ of $u$ and $k$ is the iterate step. The initial node embeddings $\textbf{h}_u^{(0)}$ are the input features of the graph, and superscripts are used to distinguish embeddings and functions at different steps of the message passing interations. 

Among the numerous GNN models, GraphSAGE (SAmple and AggreGate), is a successful variant of the vanila GCN that performs well in large graphs and, learn the embeddings in a inductive manner, generalizing for nodes that were not present during training. Based on a sampling and aggregation strategy of node neighbors that updates node embeddings through random subsets of sampled neighbors, GraphSAGE effectively reduces the dimension of the graph during training and improves the learning capacity of the model. The modified message passing equation for GraphSAGE is:

\begin{equation}~\label{eq:graphsage}
    \textbf{h}_u^{(k+1)} = \sigma\left(\textbf{W}^{(k)} \cdot \text{AGGREGATE}^{(k)}\left(\textbf{h}_u^{(k)}, \textbf{h}_v^{(k)}, \forall v \in \mathcal{N}(u)\right)\right),
\end{equation}

where $\textbf{W}^{(k)}$ weight trainable matrix, $\sigma$ is a non-linear activation function, $\mathcal{N}(u)$ is the node $u$'s neighborhood and $AGGREGATE^{(k)}$ is an aggregation function.

\section{The Graph Neural Alchemist architecture}~\label{sec:gna_architecture}

\begin{figure}[tbp]
    \centering        
    \includegraphics[width=\textwidth]{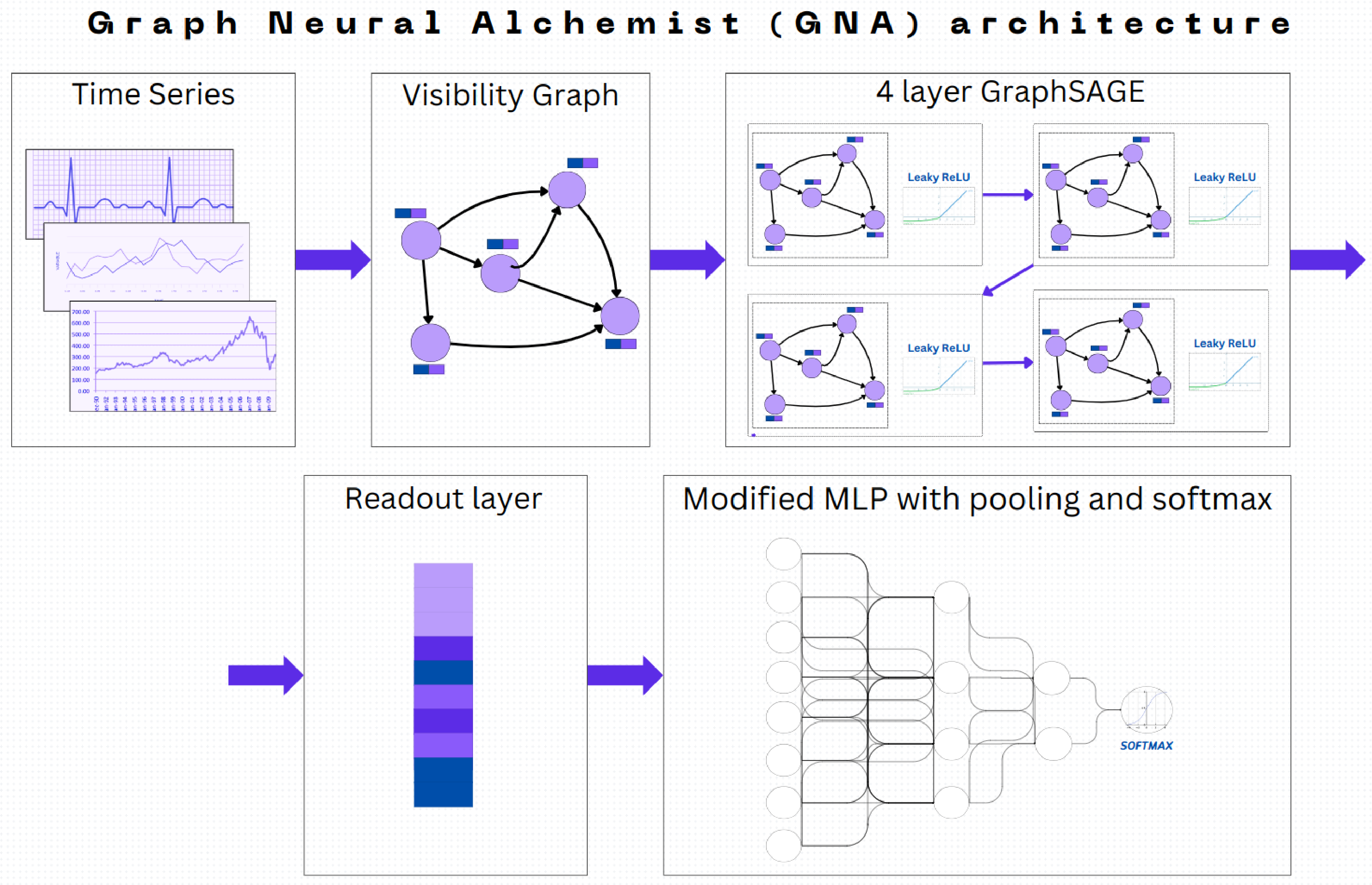}    
    \caption{The proposed GNA: a time series is fed to the architecture and each box on the right represents a fully customizable module of the architecture. Here we demonstrate the use of VG to represent the time series, followed by a 4-layer GraphSAGE network with a readout layer and a 3-layer modified MLP for classification.}~\label{fig:gna_architecture}    
\end{figure}

Designed to classify time series data represented as graphs, the Graph Neural Alchemist (GNA) architecture is fully modular and is comprised of four primary modules: the Graph Representation Module, the Graph Neural Network (GNN) Module, the Readout Layer, and the Multilayer Pooling Perceptron (MLPP) Module. Each module is highly customizable, allowing for easy substitution or even supression depending on the data domain and specific task requirements. The architecture is depicted in Figure~\ref{fig:gna_architecture}.

\subsection{Graph Representation Module}~\label{subsec:graph_representation_module}

We define a Directed Acyclic Graph (DAG) $\mathcal{G} (\mathcal{V}, \mathcal{E})$ where the nodes correspond to all the values from the time series, and the edges are established by the visibility criterion of equation~\ref{eq:visibility_criterion}. Using left-to-right directed edges and NVG to represent time series enables the model to better capture the temporal dynamics of the series as they progress over time and space.

For the initial node features, the representation encodes two key metrics: \textit{In-degree} and \textit{PageRank}. The in-degree feature quantifies the number of edges incident on a node, reflecting its local importance and connectivity. Nodes with higher in-degree are central within their local neighborhoods and the information from this feature is crucial in understanding local patterns and anomalies in the data.\\
Conversely, the PageRank feature measures the global importance of each node within the entire graph, based on the number and significance of incoming connections. This dual encoding ensures that each node captures both local and global importance, effectively representing the intrinsic relationships within the time series, and the relationships between past and future data points.

\subsection{Graph Neural Network (GNN) Module}~\label{subsec:graph_neural_network_module}

The GNA architecture employs a 4-layer GraphSAGE network, which aggregate features from sampled neighbors to learn the \textit{hidden embeddings} of the graph via the neural message passing framework (Equation~\ref{eq:graphsage}). We used the \text{Leaky ReLU} activation function between each layer to introduce non-linearity and enhance learning capabilities, defined as:

\begin{equation}~\label{eq:lealkyReLU}
    \text{Leaky ReLU}(\text{x}) = \begin{cases} 
        \text{x} & \textit{if } \space \text{x} \space \geq 0 \\
    \alpha \text{x} & \text{otherwise},
    \end{cases}
\end{equation} where $\alpha$ controls the angle of the negative slope and is set to the default value of $1e-2$.

By employing GraphSAGE, we ensure that every possible neighbor, carrying local (in-degree) and global (PageRank) information, contributes to the node's updated representation, allowing the model to capture the evolution of the time series and learn its spatio-temporal dynamics. The choice for 4 layers was to avoid oversmothing, a common issue in deep GNNs, where successive message passing iterations cause the model to converge into indistinguishable vectors for all nodes, losing the local information.

\subsection{Readout Layer}~\label{sec:readout_layer}

Following the GNN module, a readout layer aggregates the hidden features from all nodes to produce a single vector representation of the graph. This step is essential for graph classification tasks as it summarizes the learned node embeddings, encoding the information of the entire graph into a form suitable for downstream classification. Here, we use an average readout layer, which computes the weighted mean of the hidden features across all nodes of the graph.

\subsection{Multilayer Pooling Perceptron (MLPP) Module}~\label{sec:mlpp_module}

The Multilayer Pooling Perceptron (MLPP) module in our architecture performs the classification step as well as the pooling operation, a strategy borrowed from Convolutional Neural Networks (CNNs) to reduce the graph's dimensionality while preserving its essential information. While some GNN models incorporate pooling between their layers to coarsen the graph, we apply it afterwards in the MLPP module to keep the original graph structure and to enhance modularity.

Each layer in the MLPP module halves the dimension of the hidden features during the feedforward step and the final layer outputs the predicted probabilities for each class that is then passed through a \text{log softmax} activation function to obtain the logit values. To introduce non-linearity, a \text{Leaky ReLU} activation function is used between each layer. The \text{log softmax} function is defined as:

\begin{equation}~\label{eq:log_softmax}
    \text{Log Softmax}(\mathbf{h})_i = \log \left( \frac{\exp(h_i)}{\sum_{j} \exp(h_j)} \right),
\end{equation} 

where $\mathbf{h}$ is the learned embedding vector of the graph and $i$ is the index of the class.

The weights of the networks are learned using the Adam stochastic gradient descent optimizer, which optimizes the cross-entropy loss function, defined as:

\begin{equation}~\label{eq:cross_entropy_loss}
    \mathcal{L} = -\sum_{i=1}^{N} y_i \log(\hat{y}_i),
\end{equation}

where $y_i$ is the true label and $\hat{y}_i$ is the predicted probability for class $i$.

Considering the VG representation, GraphSAGE and MLPP used here, the overall time complexity of the architecture is $\text{O}(M \times (E + N))$, where $M$ is the number of time series, $E$ is the number of edges, and $N$ is the number of nodes in the graph, which is far more efficient than traditional time series classification models.

\section{Experimental Results}~\label{sec:experimental_results}

We implement our GNA architecture using PyTorch~\footnote{\url{https://pytorch.org/}} and the Deep Graph Library (DGL)~\footnote{\url{https://www.dgl.ai/}} for deep learning and graph-based computations, and the ts2vg library~\footnote{\url{https://pypi.org/project/ts2vg/}} for generating visibility graphs. For evaluation, we employed Scikit-learn~\footnote{\url{https://scikit-learn.org/stable/}} and, for the benchmark model, we used the implementation found at Sktime library~\footnote{\url{https://www.sktime.net/en/stable/index.html}}. All experiments were conducted on a server running Ubuntu 22.10, powered by an AMD Ryzen 9 5950X 16-Core Processor and equipped with a 24GB NVIDIA GeForce RTX 3090 Ti GPU. To leverage the GPU's computational power, we used CUDA Version 12.4 for accelerated deep learning operations.

The UCR Time Series Classification Archive is a well-known repository of diverse time series datasets used for classification benchmarks~\citep{UCRArchive2018}. For this study, we selected ten datasets from the UCR Archive to evaluate our GNA architecture (Table~\ref{tab:datasets}). These datasets were chosen based on their diversity in terms of the number of time series, samples, and classes, as well as their complexity and imbalance.

\begin{table}[htbp]
    \centering
    \caption{Description of the Datasets}
    \begin{tabular}{|l|l|l|l|}
    \hline
    \textbf{Dataset} & \textbf{No of Time Series} & \textbf{No of Samples} & \textbf{No of Classes} \\
    \hline
    Crop & 24000 & 46 & 24 \\
    Earthquakes & 460 & 512 & 2 \\
    ECG200 & 200 & 95 & 2 \\
    ECGFiveDays & 884 & 136 & 2 \\
    ElectricDevices & 16637 & 96 & 7 \\
    NonInvasiveFetalECGThorax1 & 3765 & 750 & 42 \\
    NonInvasiveFetalECGThorax2 & 3765 & 750 & 42 \\
    Phoneme & 2110 & 1024 & 39 \\
    Strawberry & 983 & 235 & 2 \\
    TwoLeadECG & 1162 & 82 & 2 \\
    \hline
    \end{tabular}~\label{tab:datasets}
\end{table}

Given the inherent randomness in the initialization of deep learning models, their performance can vary significantly between executions. To address this and to better assess classification performance, we devised a comprehensive strategy: we employed a random search to determine the optimal hyperparameters for our GNA model across the following choice intervals: {
    nhid: [8, 16, 32, 64, 128], 
    lr: [1e-4, 1e-3, 1e-2, 1e-1], 
    epochs: [i for i in range(50, 301, 50)], 
    batch\_size: [16, 32, 64, 128, 256]  
} over the Strawberry, Phoneme, and Crop datasets. The hyperparameters that yielded the best mean F1-score were selected for the experiments: \textbf{nhid:} 128, \textbf{lr:} 0.001, \textbf{epochs:} 250, and \textbf{batch size:} 32.

The datasets were split according to the original train-test division present in the UCR Archive. We trained and tested the GNA architecture 30 times with different random seeds, recording precision, recall, accuracy, and F1-score for each run. After the 30th run, the average of each metric was computed for analysis. We benchmarked the proposed GNA architecture against the fast ROCKET (RandOm Convolutional KErnel Transform) method, that uses a large number of randomly initialized convolutional kernels to transform time series data, extracting diverse features which are then classified using a linear classifier~\citep{dempster2020rocket}. To our experiments we used 512 kernels. Table~\ref{tab:performance_experiments} shows the results.

\begin{table}[htbp]
    \centering
    \caption{Performance metrics comparison of GNA and ROCKET classifiers on the UCR Time Series Classification Archive datasets. The results are presented as the mean and standard deviation of 30 runs.}
    \begin{adjustbox}{center}
    \begin{tabular}{cc}
        \begin{tabular}{|l|l|l|l|l|}
        \hline
        \multicolumn{5}{|c|}{\textbf{(a) Performance of GNA architecture}~\label{tab:performance_gna}} \\
        \hline
        \textit{\textbf{Dataset}} & \textit{\textbf{Precision}} & \textit{\textbf{Recall}} & \textit{\textbf{Accuracy}} & \textit{\textbf{f1-score}}\\
        \hline
        \textit{\textbf{Crop}}                       & \(0.52 \pm 0.00\) & \(0.53 \pm 0.00\) & \(0.53 \pm 0.00\) & \(0.52 \pm 0.00\)\\
        \textit{\textbf{Earthquakes}}                & \(0.37 \pm 0.00\) & \(0.50 \pm 0.00\) & \(0.75 \pm 0.00\) & \(0.43 \pm 0.00\)\\
        \textit{\textbf{ECG200}}                     & \(0.72 \pm 0.00\) & \(0.67 \pm 0.00\) & \(0.73 \pm 0.00\) & \(0.68 \pm 0.00\)\\
        \textit{\textbf{ECGFiveDays}}                & \(0.75 \pm 0.04\) & \(0.72 \pm 0.09\) & \(0.72 \pm 0.09\) & \(0.70 \pm 0.14\)\\
        \textit{\textbf{ElectricDevices}}            & \(0.63 \pm 0.01\) & \(0.60 \pm 0.01\) & \(0.68 \pm 0.01\) & \(0.61 \pm 0.01\) \\
        \textit{\textbf{NonInvasiveFetalECGThorax1}} & \(0.66 \pm 0.02\) & \(0.65 \pm 0.02\) & \(0.65 \pm 0.02\) & \(0.64 \pm 0.02\) \\
        \textit{\textbf{NonInvasiveFetalECGThorax2}} & \(0.73 \pm 0.01\) & \(0.72 \pm 0.01\) & \(0.73 \pm 0.01\) & \(0.72 \pm 0.01\) \\
        \textit{\textbf{Phoneme}}                    & \(0.16 \pm 0.01\) & \(0.16 \pm 0.01\) & \(0.26 \pm 0.01\) & \(0.15 \pm 0.01\) \\
        \textit{\textbf{Strawberry}}                 & \(0.91 \pm 0.01\) & \(0.92 \pm 0.01\) & \(0.92 \pm 0.00\) & \(0.92 \pm 0.01\)  \\
        \textit{\textbf{TwoLeadECG}}                 & \(0.86 \pm 0.00\) & \(0.86 \pm 0.00\) & \(0.86 \pm 0.00\) & \(0.86 \pm 0.00\)\\
        \hline
        \end{tabular}
        \\
        \\
        \begin{tabular}{|l|l|l|l|l|}
        \hline
        \multicolumn{5}{|c|}{\textbf{(b) Performance of ROCKET Classifier}~\label{tab:performance_rocket}} \\
        \hline
        \textit{\textbf{Dataset}} & \textit{\textbf{Precision}} & \textit{\textbf{Recall}} & \textit{\textbf{Accuracy}} & \textit{\textbf{f1-score}}\\
        \hline
        \textit{\textbf{Crop}}                       & \(0.67 \pm 0.00\) & \(0.69 \pm 0.00\) & \(0.69 \pm 0.00\) & \(0.67 \pm 0.00\) \\
        \textit{\textbf{Earthquakes}}                & \(0.37 \pm 0.00\) & \(0.50 \pm 0.00\) & \(0.75 \pm 0.00\) & \(0.43 \pm 0.00\) \\
        \textit{\textbf{ECG200}}                     & \(0.89 \pm 0.02\) & \(0.89 \pm 0.02\) & \(0.90 \pm 0.02\) & \(0.89 \pm 0.02\) \\
        \textit{\textbf{ECGFiveDays}}                & \(1.00 \pm 0.00\) & \(1.00 \pm 0.00\) & \(1.00 \pm 0.00\) & \(1.00 \pm 0.00\) \\                      
        \textit{\textbf{ElectricDevices}}            & \(0.69 \pm 0.01\) & \(0.64 \pm 0.01\) & \(0.72 \pm 0.01\) & \(0.64 \pm 0.01\) \\    
        \textit{\textbf{NonInvasiveFetalECGThorax1}} & \(0.93 \pm 0.00\) & \(0.93 \pm 0.00\) & \(0.93 \pm 0.00\) & \(0.92 \pm 0.00\) \\
        \textit{\textbf{NonInvasiveFetalECGThorax2}} & \(0.95 \pm 0.00\) & \(0.95 \pm 0.00\) & \(0.95 \pm 0.00\) & \(0.95 \pm 0.00\) \\
        \textit{\textbf{Phoneme}}                    & \(0.12 \pm 0.02\) & \(0.13 \pm 0.00\) & \(0.29 \pm 0.01\) & \(0.11 \pm 0.00\) \\
        \textit{\textbf{Strawberry}}                 & \(0.96 \pm 0.00\) & \(0.97 \pm 0.00\) & \(0.97 \pm 0.00\) & \(0.97 \pm 0.00\) \\    
        \textit{\textbf{TwoLeadECG}}                 & \(1.00 \pm 0.00\) & \(1.00 \pm 0.00\) & \(1.00 \pm 0.00\) & \(1.00 \pm 0.00\) \\
        \hline
        \end{tabular}
    \end{tabular}~\label{tab:performance_experiments}
    \end{adjustbox}
\end{table}

Regarding Earthquakes, both GNA and ROCKET perform similarly. This dataset is tricky, as it is significantly imbalanced, with only 25\% of the cases labeled as positive by domain experts. Such imbalance can often obscure the true performance of a classifier, as models may struggle to correctly identify minority classes. 

\begin{figure}[tbp]
    \centering        
        \includegraphics[width=\textwidth]{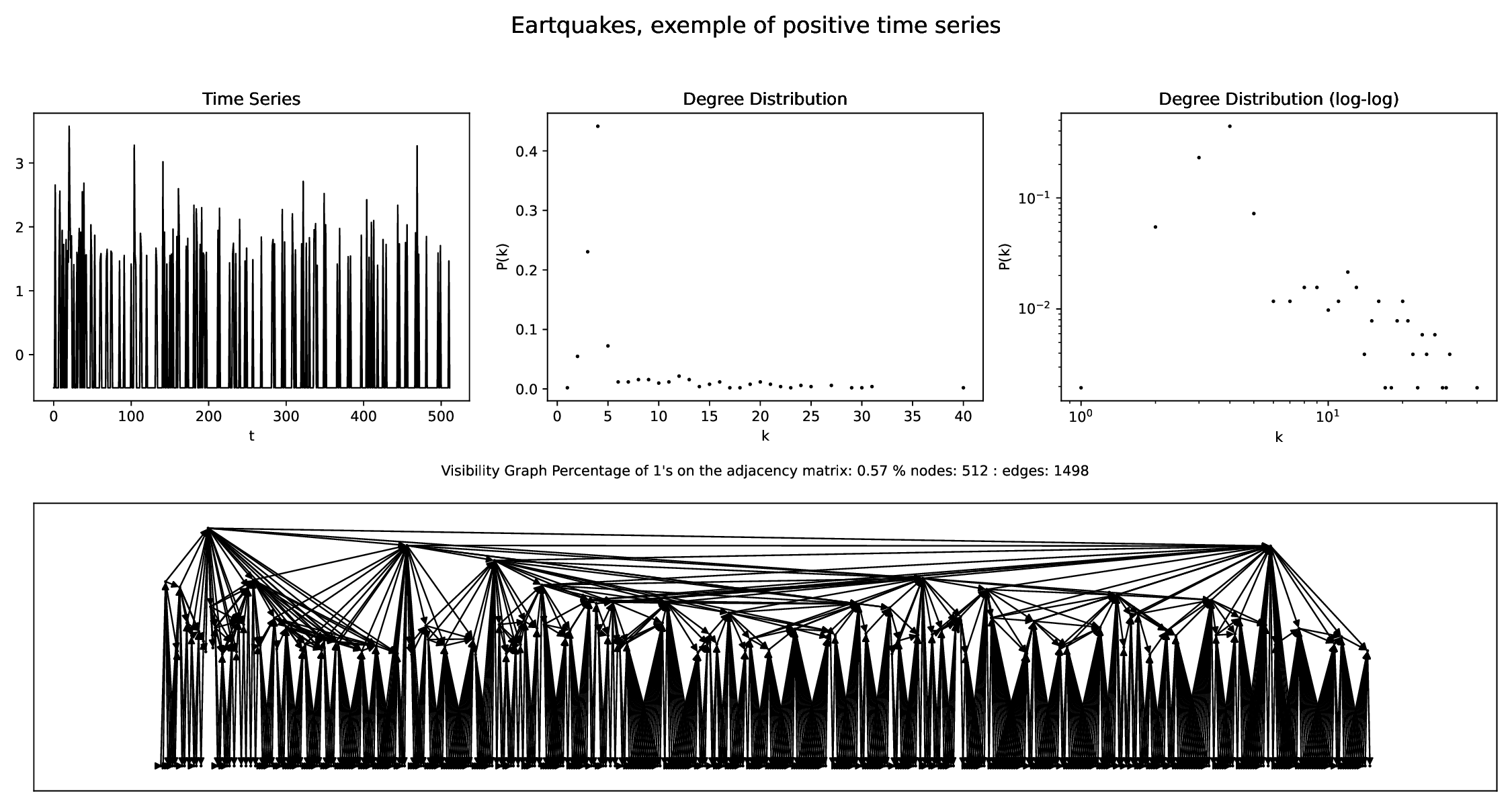}
        \caption{Spatial structure for a sample of positive-labeled time series from the Earthquakes dataset. The top left plot shows the time series, the top middle plot shows the Degree Distribution, and the top right plot shows the Degree Distribution on a log-log scale. Note how the degree distribution follows a power-law, and, in the bottom plot, how this translates into a sparse scale-free graph.}~\label{fig:earthquake}
\end{figure}

Improving on the ROCKET model in this scenario can be particularly challenging due to the model's reliance on random kernels, which may not capture the subtle spatial variations in the data.In contrast, the GNA model presents a more promising approach: as shown in figure~\ref{fig:earthquake}, the VG for an earthquake translates into a scale-free network with in-degree hubs. The poor performance observed in the GNA could be attributed to the use of a readout layer that averages the embeddings, potentially losing crucial information from these hubs. Therefore, it would be worthwhile to explore different approaches to the GNN and readout, such as attention-based methods, which could better preserve and leverage the critical information embedded in the graph.

Examining datasets with a balanced class distribution, such as ECGFiveDays and TwoLeadECG, that have have an equal 50\% distribution of each class, we note that the number of training samples is significantly lower than the test samples: only 23 training samples versus 861 test samples for ECGFiveDays and 1139 for TwoLeadECG. Our GNA model's ability to avoid overfitting in these scenarios highlights its robustness in dealing with limited training data and dataset imbalance, a challenge where the ROCKET model appears less effective. 

Examining datasets with a balanced class distribution, such as ECGFiveDays and TwoLeadECG, which have an equal 50\% distribution of each class, we note that the number of training samples is significantly lower than the test samples: only 23 training samples versus 861 test samples for ECGFiveDays and 1139 for TwoLeadECG. The GNA model's high generalization capability is evident in its performance under these conditions, where training data is scarce. This ability to perform well in scenarios with limited training data highlights a robustness that the ROCKET model appears less effective at achieving.

The Phoneme dataset stands out due to being the largest single-domain time series classification dataset in terms of signal length and to have massive amount of noise. Additionally, it has a significant imbalance in the train/test split, with only 10\% of data used for training. Despite these challenges, our GNA model outperforms ROCKET, indicating its robustness in handling noisy data, large time series, and scenarios with limited training data.

In summary, the performance of our GNA architecture varies across datasets, with its strengths and weaknesses becoming apparent in different contexts. While it struggles with imbalanced datasets and subtle spatial variations, it demonstrates robustness in handling noise, large time series, and limited training data. This suggests that our proposed architecture is well-suited to handle real-world data, notorious for being noisy and having scarce training data.

\section{Conclusion and future works}~\label{sec:conclusion}

Here we proposed the Graph Neural Alchemist (GNA) architecture, a modular framework for time series classification based on visibility graph representations and GNNs. Our architecture is designed to capture the spatio-temporal dynamics of time series data by encoding them as directed visibility graphs and learning their hidden features using a GraphSAGE network. We evaluated the GNA architecture on a variety of time series datasets from the UCR Time Series Classification Archive, comparing its performance against the ROCKET model. Our results show that the GNA architecture outperforms ROCKET in some specific cases, demonstrating its robustness in handling noisy data, large time series, and scenarios with limited training data. However, the GNA architecture struggles with imbalanced datasets and subtle spatial variations, indicating areas for improvement.

For future work, we aim to demonstrate the representation-agnostic capability of our architecture by testing it with other time series-to-graph representations, such as the Ordinal Pattern Transition Graph (OPTG), and a different classificator, such as Xgboost, to further enhance its performance. This would allow us to verify the generalization and robustness of the proposed architecture, potentially leading to a more versatile framework. Additionally, we are currently evaluating its performance on a larger real-world health dataset, to assess its applicability in clinical settings and robustness capabilities.

Furthermore, extending the classification module to evaluate the architecture on a variety of tasks, including node classification, community detection, and link prediction, would be a promising line of work. These experiments could help assess the generalization capability of the proposed architecture across different types of tasks, broadening its applicability and demonstrating its versatility.

\bibliographystyle{abbrvnat}
\bibliography{samplepaper}

\end{document}